\documentstyle[aclap,psfig]{article}  

\newcounter{example}

\newcommand{\Xb}[1] {
   \begin{list}{(\arabic{example})}{
      \setlength{\leftmargin}{5 em}
      \setlength{\rightmargin}{0in}
      \setlength{\labelwidth}{4 em}
      \setlength{\labelsep}{1 em}
      \setlength{\parsep}{0in}
      \setlength{\itemsep}{0in}
      \setlength{\listparindent}{0in}}
   \refstepcounter{example}
   \item
   \label{#1}}

\newcommand{\Xfb}[1] {\begin{figure}[htb]
   \begin{list}{(\arabic{example})}{
      \setlength{\leftmargin}{5 em}
      \setlength{\rightmargin}{0in}
      \setlength{\labelwidth}{4 em}
      \setlength{\labelsep}{1 em}
      \setlength{\parsep}{0in}
      \setlength{\itemsep}{0in}
      \setlength{\listparindent}{0in}}
   \refstepcounter{example}
   \item
   \label{#1}}

\newcommand{\Xe} {
   \end{list}}

\newcommand{\Xfe} {
   \end{list} \end{figure}}

\newcommand{\wideXb}[1] {\begin{figure*}[htb]
   \begin{list}{(\arabic{example})}{
      \setlength{\leftmargin}{5 em}
      \setlength{\rightmargin}{0in}
      \setlength{\labelwidth}{4 em}
      \setlength{\labelsep}{1 em}
      \setlength{\parsep}{0in}
      \setlength{\itemsep}{0in}
      \setlength{\listparindent}{0in}}
   \refstepcounter{example}
   \item
   \label{#1}}

\newcommand{\wideXe} {
   \end{list} \end{figure*}}

\newcommand{\Xbmult} {\begin{figure}[htb]
   \begin{list}{(\arabic{example})}{
      \setlength{\leftmargin}{5 em}
      \setlength{\rightmargin}{0in}
      \setlength{\labelwidth}{4 em}
      \setlength{\labelsep}{1 em}
      \setlength{\parsep}{0in}
      \setlength{\itemsep}{0in}
      \setlength{\listparindent}{0in}}}

\newcommand{\xref}[1] {(\ref{#1})}
\newcommand{\xaref}[2] {(\ref{#1}#2)}

\newcounter{treecount}
\newcounter{branchcount}
\newsavebox{\parentbox}
\newsavebox{\treebox}
\newsavebox{\treeboxone}
\newsavebox{\treeboxtwo}
\newsavebox{\treeboxthree}
\newsavebox{\treeboxfour}
\newsavebox{\treeboxfive}
\newsavebox{\treeboxsix}
\newsavebox{\treeboxseven}
\newsavebox{\treeboxeight}
\newsavebox{\treeboxnine}
\newsavebox{\treeboxten}
\newsavebox{\treeboxeleven}
\newsavebox{\treeboxtwelve}
\newsavebox{\treeboxthirteen}
\newsavebox{\treeboxfourteen}
\newsavebox{\treeboxfifteen}
\newsavebox{\treeboxsixteen}
\newsavebox{\treeboxseventeen}
\newsavebox{\treeboxeighteen}
\newsavebox{\treeboxnineteen}
\newsavebox{\treeboxtwenty}
\newlength{\treeoffsetone}
\newlength{\treeoffsettwo}
\newlength{\treeoffsetthree}
\newlength{\treeoffsetfour}
\newlength{\treeoffsetfive}
\newlength{\treeoffsetsix}
\newlength{\treeoffsetseven}
\newlength{\treeoffseteight}
\newlength{\treeoffsetnine}
\newlength{\treeoffsetten}
\newlength{\treeoffseteleven}
\newlength{\treeoffsettwelve}
\newlength{\treeoffsetthirteen}
\newlength{\treeoffsetfourteen}
\newlength{\treeoffsetfifteen}
\newlength{\treeoffsetsixteen}
\newlength{\treeoffsetseventeen}
\newlength{\treeoffseteighteen}
\newlength{\treeoffsetnineteen}
\newlength{\treeoffsettwenty}

\newlength{\treeshiftone}
\newlength{\treeshifttwo}
\newlength{\treeshiftthree}
\newlength{\treeshiftfour}
\newlength{\treeshiftfive}
\newlength{\treeshiftsix}
\newlength{\treeshiftseven}
\newlength{\treeshifteight}
\newlength{\treeshiftnine}
\newlength{\treeshiftten}
\newlength{\treeshifteleven}
\newlength{\treeshifttwelve}
\newlength{\treeshiftthirteen}
\newlength{\treeshiftfourteen}
\newlength{\treeshiftfifteen}
\newlength{\treeshiftsixteen}
\newlength{\treeshiftseventeen}
\newlength{\treeshifteighteen}
\newlength{\treeshiftnineteen}
\newlength{\treeshifttwenty}
\newlength{\treewidthone}
\newlength{\treewidthtwo}
\newlength{\treewidththree}
\newlength{\treewidthfour}
\newlength{\treewidthfive}
\newlength{\treewidthsix}
\newlength{\treewidthseven}
\newlength{\treewidtheight}
\newlength{\treewidthnine}
\newlength{\treewidthten}
\newlength{\treewidtheleven}
\newlength{\treewidthtwelve}
\newlength{\treewidththirteen}
\newlength{\treewidthfourteen}
\newlength{\treewidthfifteen}
\newlength{\treewidthsixteen}
\newlength{\treewidthseventeen}
\newlength{\treewidtheighteen}
\newlength{\treewidthnineteen}
\newlength{\treewidthtwenty}
\newlength{\daughteroffsetone}
\newlength{\daughteroffsettwo}
\newlength{\daughteroffsetthree}
\newlength{\daughteroffsetfour}
\newlength{\branchwidthone}
\newlength{\branchwidthtwo}
\newlength{\branchwidththree}
\newlength{\branchwidthfour}
\newlength{\parentoffset}
\newlength{\treeoffset}
\newlength{\daughteroffset}
\newlength{\branchwidth}
\newlength{\parentwidth}
\newlength{\treewidth}
\newcommand{\ontop}[1]{\begin{tabular}{c}#1\end{tabular}}
\newcommand{\poptree}{%
\ifnum\value{treecount}=0\typeout{QobiTeX warning---Tree stack underflow}\fi%
\addtocounter{treecount}{-1}%
\setlength{\treeoffsettwo}{\treeoffsetthree}%
\setlength{\treeoffsetthree}{\treeoffsetfour}%
\setlength{\treeoffsetfour}{\treeoffsetfive}%
\setlength{\treeoffsetfive}{\treeoffsetsix}%
\setlength{\treeoffsetsix}{\treeoffsetseven}%
\setlength{\treeoffsetseven}{\treeoffseteight}%
\setlength{\treeoffseteight}{\treeoffsetnine}%
\setlength{\treeoffsetnine}{\treeoffsetten}%
\setlength{\treeoffsetten}{\treeoffseteleven}%
\setlength{\treeoffseteleven}{\treeoffsettwelve}%
\setlength{\treeoffsettwelve}{\treeoffsetthirteen}%
\setlength{\treeoffsetthirteen}{\treeoffsetfourteen}%
\setlength{\treeoffsetfourteen}{\treeoffsetfifteen}%
\setlength{\treeoffsetfifteen}{\treeoffsetsixteen}%
\setlength{\treeoffsetsixteen}{\treeoffsetseventeen}%
\setlength{\treeoffsetseventeen}{\treeoffseteighteen}%
\setlength{\treeoffseteighteen}{\treeoffsetnineteen}%
\setlength{\treeoffsetnineteen}{\treeoffsettwenty}%
\setlength{\treeshifttwo}{\treeshiftthree}%
\setlength{\treeshiftthree}{\treeshiftfour}%
\setlength{\treeshiftfour}{\treeshiftfive}%
\setlength{\treeshiftfive}{\treeshiftsix}%
\setlength{\treeshiftsix}{\treeshiftseven}%
\setlength{\treeshiftseven}{\treeshifteight}%
\setlength{\treeshifteight}{\treeshiftnine}%
\setlength{\treeshiftnine}{\treeshiftten}%
\setlength{\treeshiftten}{\treeshifteleven}%
\setlength{\treeshifteleven}{\treeshifttwelve}%
\setlength{\treeshifttwelve}{\treeshiftthirteen}%
\setlength{\treeshiftthirteen}{\treeshiftfourteen}%
\setlength{\treeshiftfourteen}{\treeshiftfifteen}%
\setlength{\treeshiftfifteen}{\treeshiftsixteen}%
\setlength{\treeshiftsixteen}{\treeshiftseventeen}%
\setlength{\treeshiftseventeen}{\treeshifteighteen}%
\setlength{\treeshifteighteen}{\treeshiftnineteen}%
\setlength{\treeshiftnineteen}{\treeshifttwenty}%
\setlength{\treewidthtwo}{\treewidththree}%
\setlength{\treewidththree}{\treewidthfour}%
\setlength{\treewidthfour}{\treewidthfive}%
\setlength{\treewidthfive}{\treewidthsix}%
\setlength{\treewidthsix}{\treewidthseven}%
\setlength{\treewidthseven}{\treewidtheight}%
\setlength{\treewidtheight}{\treewidthnine}%
\setlength{\treewidthnine}{\treewidthten}%
\setlength{\treewidthten}{\treewidtheleven}%
\setlength{\treewidtheleven}{\treewidthtwelve}%
\setlength{\treewidthtwelve}{\treewidththirteen}%
\setlength{\treewidththirteen}{\treewidthfourteen}%
\setlength{\treewidthfourteen}{\treewidthfifteen}%
\setlength{\treewidthfifteen}{\treewidthsixteen}%
\setlength{\treewidthsixteen}{\treewidthseventeen}%
\setlength{\treewidthseventeen}{\treewidtheighteen}%
\setlength{\treewidtheighteen}{\treewidthnineteen}%
\setlength{\treewidthnineteen}{\treewidthtwenty}%
\sbox{\treeboxtwo}{\usebox{\treeboxthree}}%
\sbox{\treeboxthree}{\usebox{\treeboxfour}}%
\sbox{\treeboxfour}{\usebox{\treeboxfive}}%
\sbox{\treeboxfive}{\usebox{\treeboxsix}}%
\sbox{\treeboxsix}{\usebox{\treeboxseven}}%
\sbox{\treeboxseven}{\usebox{\treeboxeight}}%
\sbox{\treeboxeight}{\usebox{\treeboxnine}}%
\sbox{\treeboxnine}{\usebox{\treeboxten}}%
\sbox{\treeboxten}{\usebox{\treeboxeleven}}%
\sbox{\treeboxeleven}{\usebox{\treeboxtwelve}}%
\sbox{\treeboxtwelve}{\usebox{\treeboxthirteen}}%
\sbox{\treeboxthirteen}{\usebox{\treeboxfourteen}}%
\sbox{\treeboxfourteen}{\usebox{\treeboxfifteen}}%
\sbox{\treeboxfifteen}{\usebox{\treeboxsixteen}}%
\sbox{\treeboxsixteen}{\usebox{\treeboxseventeen}}%
\sbox{\treeboxseventeen}{\usebox{\treeboxeighteen}}%
\sbox{\treeboxeighteen}{\usebox{\treeboxnineteen}}%
\sbox{\treeboxnineteen}{\usebox{\treeboxtwenty}}}
\newcommand{\leaf}[1]{%
\ifnum\value{treecount}=20\typeout{QobiTeX warning---Tree stack overflow}\fi%
\addtocounter{treecount}{1}%
\sbox{\treeboxtwenty}{\usebox{\treeboxnineteen}}%
\sbox{\treeboxnineteen}{\usebox{\treeboxeighteen}}%
\sbox{\treeboxeighteen}{\usebox{\treeboxseventeen}}%
\sbox{\treeboxseventeen}{\usebox{\treeboxsixteen}}%
\sbox{\treeboxsixteen}{\usebox{\treeboxfifteen}}%
\sbox{\treeboxfifteen}{\usebox{\treeboxfourteen}}%
\sbox{\treeboxfourteen}{\usebox{\treeboxthirteen}}%
\sbox{\treeboxthirteen}{\usebox{\treeboxtwelve}}%
\sbox{\treeboxtwelve}{\usebox{\treeboxeleven}}%
\sbox{\treeboxeleven}{\usebox{\treeboxten}}%
\sbox{\treeboxten}{\usebox{\treeboxnine}}%
\sbox{\treeboxnine}{\usebox{\treeboxeight}}%
\sbox{\treeboxeight}{\usebox{\treeboxseven}}%
\sbox{\treeboxseven}{\usebox{\treeboxsix}}%
\sbox{\treeboxsix}{\usebox{\treeboxfive}}%
\sbox{\treeboxfive}{\usebox{\treeboxfour}}%
\sbox{\treeboxfour}{\usebox{\treeboxthree}}%
\sbox{\treeboxthree}{\usebox{\treeboxtwo}}%
\sbox{\treeboxtwo}{\usebox{\treeboxone}}%
\sbox{\treeboxone}{\ontop{#1}}%
\sbox{\treeboxone}{\raisebox{-\ht\treeboxone}{\usebox{\treeboxone}}}%
\setlength{\treeoffsettwenty}{\treeoffsetnineteen}%
\setlength{\treeoffsetnineteen}{\treeoffseteighteen}%
\setlength{\treeoffseteighteen}{\treeoffsetseventeen}%
\setlength{\treeoffsetseventeen}{\treeoffsetsixteen}%
\setlength{\treeoffsetsixteen}{\treeoffsetfifteen}%
\setlength{\treeoffsetfifteen}{\treeoffsetfourteen}%
\setlength{\treeoffsetfourteen}{\treeoffsetthirteen}%
\setlength{\treeoffsetthirteen}{\treeoffsettwelve}%
\setlength{\treeoffsettwelve}{\treeoffseteleven}%
\setlength{\treeoffseteleven}{\treeoffsetten}%
\setlength{\treeoffsetten}{\treeoffsetnine}%
\setlength{\treeoffsetnine}{\treeoffseteight}%
\setlength{\treeoffseteight}{\treeoffsetseven}%
\setlength{\treeoffsetseven}{\treeoffsetsix}%
\setlength{\treeoffsetsix}{\treeoffsetfive}%
\setlength{\treeoffsetfive}{\treeoffsetfour}%
\setlength{\treeoffsetfour}{\treeoffsetthree}%
\setlength{\treeoffsetthree}{\treeoffsettwo}%
\setlength{\treeoffsettwo}{\treeoffsetone}%
\setlength{\treeoffsetone}{0.5\wd\treeboxone}%
\setlength{\treeshifttwenty}{\treeshiftnineteen}%
\setlength{\treeshiftnineteen}{\treeshifteighteen}%
\setlength{\treeshifteighteen}{\treeshiftseventeen}%
\setlength{\treeshiftseventeen}{\treeshiftsixteen}%
\setlength{\treeshiftsixteen}{\treeshiftfifteen}%
\setlength{\treeshiftfifteen}{\treeshiftfourteen}%
\setlength{\treeshiftfourteen}{\treeshiftthirteen}%
\setlength{\treeshiftthirteen}{\treeshifttwelve}%
\setlength{\treeshifttwelve}{\treeshifteleven}%
\setlength{\treeshifteleven}{\treeshiftten}%
\setlength{\treeshiftten}{\treeshiftnine}%
\setlength{\treeshiftnine}{\treeshifteight}%
\setlength{\treeshifteight}{\treeshiftseven}%
\setlength{\treeshiftseven}{\treeshiftsix}%
\setlength{\treeshiftsix}{\treeshiftfive}%
\setlength{\treeshiftfive}{\treeshiftfour}%
\setlength{\treeshiftfour}{\treeshiftthree}%
\setlength{\treeshiftthree}{\treeshifttwo}%
\setlength{\treeshifttwo}{\treeshiftone}%
\setlength{\treeshiftone}{0pt}%
\setlength{\treewidthtwenty}{\treewidthnineteen}%
\setlength{\treewidthnineteen}{\treewidtheighteen}%
\setlength{\treewidtheighteen}{\treewidthseventeen}%
\setlength{\treewidthseventeen}{\treewidthsixteen}%
\setlength{\treewidthsixteen}{\treewidthfifteen}%
\setlength{\treewidthfifteen}{\treewidthfourteen}%
\setlength{\treewidthfourteen}{\treewidththirteen}%
\setlength{\treewidththirteen}{\treewidthtwelve}%
\setlength{\treewidthtwelve}{\treewidtheleven}%
\setlength{\treewidtheleven}{\treewidthten}%
\setlength{\treewidthten}{\treewidthnine}%
\setlength{\treewidthnine}{\treewidtheight}%
\setlength{\treewidtheight}{\treewidthseven}%
\setlength{\treewidthseven}{\treewidthsix}%
\setlength{\treewidthsix}{\treewidthfive}%
\setlength{\treewidthfive}{\treewidthfour}%
\setlength{\treewidthfour}{\treewidththree}%
\setlength{\treewidththree}{\treewidthtwo}%
\setlength{\treewidthtwo}{\treewidthone}%
\setlength{\treewidthone}{\wd\treeboxone}}
\newcommand{\branch}[2]{%
\setcounter{branchcount}{#1}%
\ifnum\value{branchcount}=1\sbox{\parentbox}{\ontop{#2}}%
\setlength{\parentoffset}{\treeoffsetone}%
\addtolength{\parentoffset}{-0.5\wd\parentbox}%
\setlength{\daughteroffset}{0in}%
\ifdim\parentoffset<0in%
\setlength{\daughteroffset}{-\parentoffset}%
\setlength{\parentoffset}{0in}\fi%
\setlength{\parentwidth}{\parentoffset}%
\addtolength{\parentwidth}{\wd\parentbox}%
\setlength{\treeoffset}{\daughteroffset}%
\addtolength{\treeoffset}{\treeoffsetone}%
\setlength{\treewidth}{\wd\treeboxone}%
\addtolength{\treewidth}{\daughteroffset}%
\ifdim\treewidth<\parentwidth\setlength{\treewidth}{\parentwidth}\fi%
\sbox{\treebox}{\begin{minipage}{\treewidth}%
\begin{flushleft}%
\hspace*{\parentoffset}\usebox{\parentbox}\\
{\setlength{\unitlength}{2ex}%
\hspace*{\treeoffset}\begin{picture}(0,1)%
\put(0,0){\line(0,1){1}}%
\end{picture}}\\
\vspace{-\baselineskip}
\hspace*{\daughteroffset}%
\raisebox{-\ht\treeboxone}{\usebox{\treeboxone}}%
\end{flushleft}%
\end{minipage}}%
\setlength{\treeoffsetone}{\parentoffset}%
\addtolength{\treeoffsetone}{0.5\wd\parentbox}%
\setlength{\treeshiftone}{0pt}%
\setlength{\treewidthone}{\treewidth}%
\sbox{\treeboxone}{\usebox{\treebox}}%
\else\ifnum\value{branchcount}=2\sbox{\parentbox}{\ontop{#2}}%
\setlength{\branchwidthone}{\treewidthtwo}%
\addtolength{\branchwidthone}{\treeoffsetone}%
\addtolength{\branchwidthone}{-\treeshiftone}%
\addtolength{\branchwidthone}{-\treeoffsettwo}%
\setlength{\branchwidth}{\branchwidthone}%
\setlength{\daughteroffsetone}{\branchwidth}%
\addtolength{\daughteroffsetone}{-\branchwidthone}%
\addtolength{\daughteroffsetone}{-\treeshiftone}%
\setlength{\parentoffset}{-0.5\wd\parentbox}%
\addtolength{\parentoffset}{\treeoffsettwo}%
\addtolength{\parentoffset}{0.5\branchwidth}%
\setlength{\daughteroffset}{0in}%
\ifdim\parentoffset<0in%
\setlength{\daughteroffset}{-\parentoffset}%
\setlength{\parentoffset}{0in}\fi%
\setlength{\parentwidth}{\parentoffset}%
\addtolength{\parentwidth}{\wd\parentbox}%
\setlength{\treeoffset}{\daughteroffset}%
\addtolength{\treeoffset}{\treeoffsettwo}%
\setlength{\treewidth}{\wd\treeboxone}%
\addtolength{\treewidth}{\daughteroffsetone}%
\addtolength{\treewidth}{\treewidthtwo}%
\addtolength{\treewidth}{\daughteroffset}%
\ifdim\treewidth<\parentwidth\setlength{\treewidth}{\parentwidth}\fi%
\sbox{\treebox}{\begin{minipage}{\treewidth}%
\begin{flushleft}%
\hspace*{\parentoffset}\usebox{\parentbox}\\
{\setlength{\unitlength}{0.5\branchwidth}%
\hspace*{\treeoffset}\begin{picture}(2,0.5)%
\put(0,0){\line(2,1){1}}%
\put(2,0){\line(-2,1){1}}%
\end{picture}}\\
\vspace{-\baselineskip}
\hspace*{\daughteroffset}%
\makebox[\treewidthtwo][l]%
{\raisebox{-\ht\treeboxtwo}{\usebox{\treeboxtwo}}}%
\hspace*{\daughteroffsetone}%
\raisebox{-\ht\treeboxone}{\usebox{\treeboxone}}%
\end{flushleft}%
\end{minipage}}%
\setlength{\treeoffsetone}{\parentoffset}%
\addtolength{\treeoffsetone}{0.5\wd\parentbox}%
\setlength{\treeshiftone}{0pt}%
\setlength{\treewidthone}{\treewidth}%
\sbox{\treeboxone}{\usebox{\treebox}}\poptree%
\else\ifnum\value{branchcount}=3\sbox{\parentbox}{\ontop{#2}}%
\setlength{\branchwidthone}{\treewidthtwo}%
\addtolength{\branchwidthone}{\treeoffsetone}%
\addtolength{\branchwidthone}{-\treeshiftone}%
\addtolength{\branchwidthone}{-\treeoffsettwo}%
\setlength{\branchwidthtwo}{\treewidththree}%
\addtolength{\branchwidthtwo}{\treeoffsettwo}%
\addtolength{\branchwidthtwo}{-\treeshifttwo}%
\addtolength{\branchwidthtwo}{-\treeoffsetthree}%
\setlength{\branchwidth}{\branchwidthone}%
\ifdim\branchwidthtwo>\branchwidth%
\setlength{\branchwidth}{\branchwidthtwo}\fi%
\setlength{\daughteroffsetone}{\branchwidth}%
\addtolength{\daughteroffsetone}{-\branchwidthone}%
\addtolength{\daughteroffsetone}{-\treeshiftone}%
\setlength{\daughteroffsettwo}{\branchwidth}%
\addtolength{\daughteroffsettwo}{-\branchwidthtwo}%
\addtolength{\daughteroffsettwo}{-\treeshifttwo}%
\setlength{\parentoffset}{-0.5\wd\parentbox}%
\addtolength{\parentoffset}{\treeoffsetthree}%
\addtolength{\parentoffset}{\branchwidth}%
\setlength{\daughteroffset}{0in}%
\ifdim\parentoffset<0in%
\setlength{\daughteroffset}{-\parentoffset}%
\setlength{\parentoffset}{0in}\fi%
\setlength{\parentwidth}{\parentoffset}%
\addtolength{\parentwidth}{\wd\parentbox}%
\setlength{\treeoffset}{\daughteroffset}%
\addtolength{\treeoffset}{\treeoffsetthree}%
\setlength{\treewidth}{\wd\treeboxone}%
\addtolength{\treewidth}{\daughteroffsetone}%
\addtolength{\treewidth}{\treewidthtwo}%
\addtolength{\treewidth}{\daughteroffsettwo}%
\addtolength{\treewidth}{\treewidththree}%
\addtolength{\treewidth}{\daughteroffset}%
\ifdim\treewidth<\parentwidth\setlength{\treewidth}{\parentwidth}\fi%
\sbox{\treebox}{\begin{minipage}{\treewidth}%
\begin{flushleft}%
\hspace*{\parentoffset}\usebox{\parentbox}\\
{\setlength{\unitlength}{0.5\branchwidth}%
\hspace*{\treeoffset}\begin{picture}(4,1)%
\put(0,0){\line(2,1){2}}%
\put(2,0){\line(0,1){1}}%
\put(4,0){\line(-2,1){2}}%
\end{picture}}\\
\vspace{-\baselineskip}
\hspace*{\daughteroffset}%
\makebox[\treewidththree][l]%
{\raisebox{-\ht\treeboxthree}{\usebox{\treeboxthree}}}%
\hspace*{\daughteroffsettwo}%
\makebox[\treewidthtwo][l]%
{\raisebox{-\ht\treeboxtwo}{\usebox{\treeboxtwo}}}%
\hspace*{\daughteroffsetone}%
\raisebox{-\ht\treeboxone}{\usebox{\treeboxone}}%
\end{flushleft}%
\end{minipage}}%
\setlength{\treeoffsetone}{\parentoffset}%
\addtolength{\treeoffsetone}{0.5\wd\parentbox}%
\setlength{\treeshiftone}{0pt}%
\setlength{\treewidthone}{\treewidth}%
\sbox{\treeboxone}{\usebox{\treebox}}\poptree\poptree%
\else\ifnum\value{branchcount}=4\sbox{\parentbox}{\ontop{#2}}%
\setlength{\branchwidthone}{\treewidthtwo}%
\addtolength{\branchwidthone}{\treeoffsetone}%
\addtolength{\branchwidthone}{-\treeshiftone}%
\addtolength{\branchwidthone}{-\treeoffsettwo}%
\setlength{\branchwidthtwo}{\treewidththree}%
\addtolength{\branchwidthtwo}{\treeoffsettwo}%
\addtolength{\branchwidthtwo}{-\treeshifttwo}%
\addtolength{\branchwidthtwo}{-\treeoffsetthree}%
\setlength{\branchwidththree}{\treewidthfour}%
\addtolength{\branchwidththree}{\treeoffsetthree}%
\addtolength{\branchwidththree}{-\treeshiftthree}%
\addtolength{\branchwidththree}{-\treeoffsetfour}%
\setlength{\branchwidth}{\branchwidthone}%
\ifdim\branchwidthtwo>\branchwidth%
\setlength{\branchwidth}{\branchwidthtwo}\fi%
\ifdim\branchwidththree>\branchwidth%
\setlength{\branchwidth}{\branchwidththree}\fi%
\setlength{\daughteroffsetone}{\branchwidth}%
\addtolength{\daughteroffsetone}{-\branchwidthone}%
\addtolength{\daughteroffsetone}{-\treeshiftone}%
\setlength{\daughteroffsettwo}{\branchwidth}%
\addtolength{\daughteroffsettwo}{-\branchwidthtwo}%
\addtolength{\daughteroffsettwo}{-\treeshifttwo}%
\setlength{\daughteroffsetthree}{\branchwidth}%
\addtolength{\daughteroffsetthree}{-\branchwidththree}%
\addtolength{\daughteroffsetthree}{-\treeshiftthree}%
\setlength{\parentoffset}{-0.5\wd\parentbox}%
\addtolength{\parentoffset}{\treeoffsetfour}%
\addtolength{\parentoffset}{1.5\branchwidth}%
\setlength{\daughteroffset}{0in}%
\ifdim\parentoffset<0in%
\setlength{\daughteroffset}{-\parentoffset}%
\setlength{\parentoffset}{0in}\fi%
\setlength{\parentwidth}{\parentoffset}%
\addtolength{\parentwidth}{\wd\parentbox}%
\setlength{\treeoffset}{\daughteroffset}%
\addtolength{\treeoffset}{\treeoffsetfour}%
\setlength{\treewidth}{\wd\treeboxone}%
\addtolength{\treewidth}{\daughteroffsetone}%
\addtolength{\treewidth}{\treewidthtwo}%
\addtolength{\treewidth}{\daughteroffsettwo}%
\addtolength{\treewidth}{\treewidththree}%
\addtolength{\treewidth}{\daughteroffsetthree}%
\addtolength{\treewidth}{\treewidthfour}%
\addtolength{\treewidth}{\daughteroffset}%
\ifdim\treewidth<\parentwidth\setlength{\treewidth}{\parentwidth}\fi%
\sbox{\treebox}{\begin{minipage}{\treewidth}%
\begin{flushleft}%
\hspace*{\parentoffset}\usebox{\parentbox}\\
{\setlength{\unitlength}{0.5\branchwidth}%
\hspace*{\treeoffset}\begin{picture}(6,1)%
\put(0,0){\line(3,1){3}}%
\put(2,0){\line(1,1){1}}%
\put(4,0){\line(-1,1){1}}%
\put(6,0){\line(-3,1){3}}%
\end{picture}}\\
\vspace{-\baselineskip}
\hspace*{\daughteroffset}%
\makebox[\treewidthfour][l]%
{\raisebox{-\ht\treeboxfour}{\usebox{\treeboxfour}}}%
\hspace*{\daughteroffsetthree}%
\makebox[\treewidththree][l]%
{\raisebox{-\ht\treeboxthree}{\usebox{\treeboxthree}}}%
\hspace*{\daughteroffsettwo}%
\makebox[\treewidthtwo][l]%
{\raisebox{-\ht\treeboxtwo}{\usebox{\treeboxtwo}}}%
\hspace*{\daughteroffsetone}%
\raisebox{-\ht\treeboxone}{\usebox{\treeboxone}}%
\end{flushleft}%
\end{minipage}}%
\setlength{\treeoffsetone}{\parentoffset}%
\addtolength{\treeoffsetone}{0.5\wd\parentbox}%
\setlength{\treeshiftone}{0pt}%
\setlength{\treewidthone}{\treewidth}%
\sbox{\treeboxone}{\usebox{\treebox}}\poptree\poptree\poptree%
\else\ifnum\value{branchcount}=5\sbox{\parentbox}{\ontop{#2}}%
\setlength{\branchwidthone}{\treewidthtwo}%
\addtolength{\branchwidthone}{\treeoffsetone}%
\addtolength{\branchwidthone}{-\treeshiftone}%
\addtolength{\branchwidthone}{-\treeoffsettwo}%
\setlength{\branchwidthtwo}{\treewidththree}%
\addtolength{\branchwidthtwo}{\treeoffsettwo}%
\addtolength{\branchwidthtwo}{-\treeshifttwo}%
\addtolength{\branchwidthtwo}{-\treeoffsetthree}%
\setlength{\branchwidththree}{\treewidthfour}%
\addtolength{\branchwidththree}{\treeoffsetthree}%
\addtolength{\branchwidththree}{-\treeshiftthree}%
\addtolength{\branchwidththree}{-\treeoffsetfour}%
\setlength{\branchwidthfour}{\treewidthfive}%
\addtolength{\branchwidthfour}{\treeoffsetfour}%
\addtolength{\branchwidthfour}{-\treeshiftfour}%
\addtolength{\branchwidthfour}{-\treeoffsetfive}%
\setlength{\branchwidth}{\branchwidthone}%
\ifdim\branchwidthtwo>\branchwidth%
\setlength{\branchwidth}{\branchwidthtwo}\fi%
\ifdim\branchwidththree>\branchwidth%
\setlength{\branchwidth}{\branchwidththree}\fi%
\ifdim\branchwidthfour>\branchwidth%
\setlength{\branchwidth}{\branchwidthfour}\fi%
\setlength{\daughteroffsetone}{\branchwidth}%
\addtolength{\daughteroffsetone}{-\branchwidthone}%
\addtolength{\daughteroffsetone}{-\treeshiftone}%
\setlength{\daughteroffsettwo}{\branchwidth}%
\addtolength{\daughteroffsettwo}{-\branchwidthtwo}%
\addtolength{\daughteroffsettwo}{-\treeshifttwo}%
\setlength{\daughteroffsetthree}{\branchwidth}%
\addtolength{\daughteroffsetthree}{-\branchwidththree}%
\addtolength{\daughteroffsetthree}{-\treeshiftthree}%
\setlength{\daughteroffsetfour}{\branchwidth}%
\addtolength{\daughteroffsetfour}{-\branchwidthfour}%
\addtolength{\daughteroffsetfour}{-\treeshiftfour}%
\setlength{\parentoffset}{-0.5\wd\parentbox}%
\addtolength{\parentoffset}{\treeoffsetfive}%
\addtolength{\parentoffset}{2\branchwidth}%
\setlength{\daughteroffset}{0in}%
\ifdim\parentoffset<0in%
\setlength{\daughteroffset}{-\parentoffset}%
\setlength{\parentoffset}{0in}\fi%
\setlength{\parentwidth}{\parentoffset}%
\addtolength{\parentwidth}{\wd\parentbox}%
\setlength{\treeoffset}{\daughteroffset}%
\addtolength{\treeoffset}{\treeoffsetfive}%
\setlength{\treewidth}{\wd\treeboxone}%
\addtolength{\treewidth}{\daughteroffsetone}%
\addtolength{\treewidth}{\treewidthtwo}%
\addtolength{\treewidth}{\daughteroffsettwo}%
\addtolength{\treewidth}{\treewidththree}%
\addtolength{\treewidth}{\daughteroffsetthree}%
\addtolength{\treewidth}{\treewidthfour}%
\addtolength{\treewidth}{\daughteroffsetfour}%
\addtolength{\treewidth}{\treewidthfive}%
\addtolength{\treewidth}{\daughteroffset}%
\ifdim\treewidth<\parentwidth\setlength{\treewidth}{\parentwidth}\fi%
\sbox{\treebox}{\begin{minipage}{\treewidth}%
\begin{flushleft}%
\hspace*{\parentoffset}\usebox{\parentbox}\\
{\setlength{\unitlength}{0.5\branchwidth}%
\hspace*{\treeoffset}\begin{picture}(8,1)%
\put(0,0){\line(4,1){4}}%
\put(2,0){\line(2,1){2}}%
\put(4,0){\line(0,1){1}}%
\put(6,0){\line(-2,1){2}}%
\put(8,0){\line(-4,1){4}}%
\end{picture}}\\
\vspace{-\baselineskip}
\hspace*{\daughteroffset}%
\makebox[\treewidthfive][l]%
{\raisebox{-\ht\treeboxfour}{\usebox{\treeboxfive}}}%
\hspace*{\daughteroffsetfour}%
\makebox[\treewidthfour][l]%
{\raisebox{-\ht\treeboxfour}{\usebox{\treeboxfour}}}%
\hspace*{\daughteroffsetthree}%
\makebox[\treewidththree][l]%
{\raisebox{-\ht\treeboxthree}{\usebox{\treeboxthree}}}%
\hspace*{\daughteroffsettwo}%
\makebox[\treewidthtwo][l]%
{\raisebox{-\ht\treeboxtwo}{\usebox{\treeboxtwo}}}%
\hspace*{\daughteroffsetone}%
\raisebox{-\ht\treeboxone}{\usebox{\treeboxone}}%
\end{flushleft}%
\end{minipage}}%
\setlength{\treeoffsetone}{\parentoffset}%
\addtolength{\treeoffsetone}{0.5\wd\parentbox}%
\setlength{\treeshiftone}{0pt}%
\setlength{\treewidthone}{\treewidth}%
\sbox{\treeboxone}{\usebox{\treebox}}\poptree\poptree\poptree\poptree%
\else\typeout{QobiTeX warning--- Can't handle #1 branching}\fi\fi\fi\fi\fi}
\newcommand{\tree}{%
\usebox{\treeboxone}
\setlength{\treeoffsetone}{\treeoffsettwo}%
\sbox{\treeboxone}{\usebox{\treeboxtwo}}%
\poptree}

\newcommand{\ssc}[1]{{\small {\bf #1}}}

\author{Barbara Di Eugenio\thanks{Learning Research \& Development
Center} \hspace*{1cm} Johanna D. Moore\thanks{Computer Science Department, and
Learning Research \& Development Center} \hspace*{1cm}  Massimo
Paolucci\thanks{Intelligent Systems Program}\\
University of Pittsburgh\\
Pittsburgh, PA 15260, USA \\
{\tt \{dieugeni,jmoore,paolucci\}@cs.pitt.edu}}
\title{\submitted{{\em Proceedings of the 35th
Conference of the Association for Computational Linguistics (ACL97),
Madrid, Spain, July 1997}}Learning Features that Predict Cue Usage}

\makeatletter
 \def\submitted#1{\setbox\@tempboxa\vbox{\normalsize \tt \raggedright
    #1 \\ \hbox{}}
    \vspace{-1.5 cm} \usebox\@tempboxa \\
    \vspace{-\ht\@tempboxa} \vspace{1.5 cm}}
\makeatother

\begin{document}
\maketitle

\bibliographystyle{fullname}

\begin{abstract}
Our goal is to identify the features that predict the
occurrence and placement of discourse cues in tutorial explanations
in order to aid in the automatic generation of explanations.
Previous attempts to devise rules for text generation were
based on intuition or small numbers of constructed examples.
We apply a machine learning program, C4.5,
to induce decision trees for cue occurrence and placement
from a corpus of data coded for a variety of features previously
thought to affect cue usage.  Our experiments enable us to
identify the  features with most predictive power,
and show that machine learning can be used to induce 
decision trees useful for text generation.
\end{abstract} 

\section{Introduction}

{\em Discourse cues} are words or phrases, such as {\em because},
{\em first}, and {\em although}, that mark structural and
semantic relationships between discourse entities. They play a crucial
role in many discourse processing tasks, including plan recognition
\cite{LitmanCogSci}, text comprehension
\cite{CohenColing84,Hobbs85,MannThompsonDP,ReichmanAIJ}, and anaphora
resolution \cite{GroszSidnerCL}.  Moreover, research in reading
comprehension indicates that felicitous use of cues improves
comprehension and recall \cite{GoldmanTR}, but that their indiscriminate use
may have detrimental effects on recall
\cite{Millisetal}.

Our goal is to identify general strategies for cue usage that can be
implemented for automatic text generation.  From the generation
perspective, cue usage consists of three  distinct, but
interrelated problems: (1)  {\em occurrence}: whether or not to
include a  cue in the generated text, (2)  {\em
placement}: where the cue should be placed in the text, and (3) 
{\em selection}: what lexical item(s) should be used.

Prior work in text generation has focused on cue selection
\cite{McKeownElhadadNLGW88Book,ElhadadConn}, or on the relation
between cue occurrence and placement and specific  rhetorical structures
\cite{RosnerNLGW6,ScottEWNLG90,VanderLindenCL}.
Other hypotheses about cue usage derive from work on 
discourse coherence and structure.
Previous research
\cite{Hobbs85,GroszSidnerCL,SchiffrinBook,MannRSTTEXT,ElhadadConn}, which has been largely
descriptive, suggests factors such as  structural features
of the discourse (e.g., level of embedding and segment complexity),
intentional and informational relations in that structure, ordering of
relata, and syntactic form of discourse constituents.

Moser and Moore \shortcite{MoserMooreACL95,MoserMooreLanguage} coded a
corpus of naturally occurring tutorial explanations for the range of
features identified in prior work.  Because they were also interested
in the contrast between occurrence and non-occurrence of cues, they
exhaustively coded for all of the factors thought to contribute to cue
usage in all of the text.  From their study, Moser and Moore
identified several interesting correlations between particular
features and specific aspects of cue usage, and were able to test
specific hypotheses from the literature that were based on constructed
examples.

In this paper, we focus on cue occurrence and placement, and present
an empirical study of the hypotheses provided by previous research,
which have never been systematically evaluated with naturally
occurring data.  We use a machine learning program, C4.5
\cite{quinlan93},  on the tagged corpus of
Moser and Moore to induce decision trees. The number of coded features 
and their interactions makes the manual construction of rules
that predict cue occurrence and placement an intractable task.

Our results largely confirm the suggestions from the literature, and
clarify them by highlighting the most influential features for a
particular task.  Discourse structure, in terms of both segment
structure and levels of embedding, affects cue occurrence the most;
intentional relations also play an important role. For cue placement,
the most important factors are syntactic structure and segment
complexity.

The paper is organized as follows. In Section~\ref{related-work} we
discuss   previous research in more detail.
Section~\ref{rda} provides an overview of Moser and Moore's coding scheme.
In Section~\ref{learning} we present  our learning experiments, and in 
Section~\ref{conclude} we discuss our results  and conclude.

\section{Related Work}
\label{related-work}

McKeown and Elhadad \shortcite{McKeownElhadadNLGW88Book,ElhadadConn} studied
several connectives (e.g., {\em but}, {\em since}, {\em because}), and 
include many insightful hypotheses about cue selection; their
observation that the distinction between {\em but} and {\em although}
depends on the {\em point} of the move is related to the notion
of {\em core} discussed below.  However, they do not address the
problem of cue occurrence.

Other researchers \cite{RosnerNLGW6,ScottEWNLG90} are concerned with
generating text from ``RST trees'', hierarchical structures where leaf
nodes contain content and internal nodes indicate the {\em rhetorical
relations\/}, as defined in Rhetorical Structure Theory (RST)
\cite{MannRSTTEXT}, that exist between subtrees. They proposed
heuristics for including and  choosing cues based
on the rhetorical relation between spans of text, the order of the
relata, and the complexity of the related text spans.  
However, \cite{ScottEWNLG90} was based on a small number of constructed examples, and
\cite{RosnerNLGW6} focused on a small number of RST relations.

\cite{litman-jair96} and  \cite{siegel-aaai94} have applied machine
learning to disambiguate between the {\em discourse} and {\em
sentential} usages of cues; however, they do not consider the issues
of occurrence and placement, and approach the problem from the point
of view of interpretation.  We closely follow the approach in
\cite{litman-jair96} in two ways. First, we use C4.5. Second, we
experiment first with each feature individually, and then with
``interesting'' subsets of features.

\section{Relational Discourse Analysis}

\label{rda}

This section briefly describes  {\em Relational Discourse Analysis} ({\em
RDA}) \cite{MoserMooreRDATR}, the coding scheme used to tag the data for 
our machine learning experiments.\footnote{For more detail about the RDA
coding scheme see
\cite{MoserMooreACL95,MoserMooreLanguage}.}

RDA is a scheme devised for analyzing tutorial explanations in the
domain of electronics troubleshooting.  It synthesizes ideas from 
\cite{GroszSidnerCL} and from RST 
\cite{MannRSTTEXT}.  Coders use RDA to exhaustively
analyze each explanation in the corpus, i.e., every word in each explanation
belongs to exactly one element in the analysis.  
An explanation may consist of multiple {\em segments}.
Each segment originates with an intention of the speaker.
Segments are internally structured and consist of a {\em core}, i.e., that
element that most directly expresses the segment purpose, and any number of
{\em contributors}, i.e. the remaining constituents. 
For each contributor, one analyzes its relation to the
core from an intentional perspective, i.e., how it is intended to
support the core, and from an informational perspective, i.e., how its
content relates to that of the core.  The set of intentional relations
in RDA is a modification of the presentational relations of RST, while
informational relations are similar to the subject matter relations in
RST.  Each segment constituent, both core and contributors, may itself
be a segment with a {\em core:contributor} structure. In some cases the core
is not explicit. This is often the case with the whole tutor's explanation, 
since its purpose is to answer the  student's explicit question.

\begin{small}
\wideXb{seg2}
\begin{tabular}{llp{4.5in}}
\ssc{Although}&A.&you know that part1 is good,\\
&B.&you should eliminate part2\\
&&before troubleshooting inside part3. \\
\ssc{This is}&&\\\ssc{because}&C.&~
\begin{tabular}{llp{3.15in}}
&1.&part2 is moved frequently \\
\ssc{and thus}&2.&is more susceptible to damage than part3.\end{tabular}\\
\ssc{Also,}&D.&it is more work to open up part3 for testing \\
\ssc{and}&E.&the process of opening drawers and extending cards in part3\\
&&may induce problems which did not already exist.  
\end{tabular}
\wideXe
\end{small}

As an example of the application of RDA, consider the partial tutor
explanation in~\xref{seg2}\footnote{To make the example more
intelligible, we replaced references to parts of the circuit with the
labels {\em part1}, {\em part2} and {\em part3}.}.  The purpose of
this segment is to inform the student that she made the strategy error
of testing inside part3 too soon.  The constituent that makes the
purpose obvious, in this case~\xaref{seg2}{-B}, is the core of the
segment.  The other constituents help to serve the segment purpose by
contributing to it. \xaref{seg2}{-C} is an example of
subsegment with its own {\em core:contributor} structure; its purpose
is to give a reason for testing part2 first.

\begin{figure*}
\begin{center}
\begin{small}
\leaf{\ssc{Although}}
\leaf{A}
\branch{2}{\em concede\\ \em criterion:act}
\leaf{(This is)\\ \ssc{because}}
\leaf{C.1}
\leaf{\ssc{and}\\ \ssc{thus}}
\branch{2}{\em convince\\ \em cause:effect}
\branch{1}{C.2}
\branch{2}{\em convince\\ \em act:reason}
\leaf{\ssc{Also}}
\leaf{D}
\branch{2}{\em convince\\ \em act:reason}
\leaf{\ssc{and}}
\leaf{E}
\branch{2}{\em convince\\ \em act:reason}
\branch{4}{B. you should eliminate part2\\
before troubleshooting inside part3}
\tree
\end{small}
\end{center}
\caption{The RDA analysis of {\protect \xref{seg2}}}
\label{relexpl}
\end{figure*}

The RDA analysis of \xref{seg2} is shown schematically in
Figure~\ref{relexpl}.  The core is depicted as the mother of all the
relations it participates in.  Each relation node is labeled with both
its intentional and informational relation, with the order of relata
in the label indicating the linear order in the discourse.  Each
relation node has up to two daughters: the cue, if any, and the
contributor, in the order they appear in the discourse.

Coders analyze each explanation in the corpus and enter their analyses into a
database.  
The corpus consists of 854 clauses comprising 668 segments, for a
total of 780 relations.  Table~\ref{distribution} summarizes the
distribution of different relations, and the number of cued relations
in each category.  Joints are segments comprising more than one core, but
no contributor; clusters are multiunit structures with no recognizable
{\em core:contributor} relation. \xaref{seg2}{-B} is a cluster
composed of two units (the two clauses), related only at the
informational level by a temporal relation.  Both clauses describe
actions, with the first action description  embedded in a {\em
matrix} (``You should'').
Cues are much more likely to occur in clusters, where only
informational relations occur, than in {\em core:contributor} structures,
where intentional and informational relations co-occur ({\em
$\chi^2$~=~33.367, p~$<$.001, df~=~1}).  In the following, we will not
discuss joints and clusters any further.

\begin{table*}
\centering
\begin{tabular}{||l||r|r||}\hline \hline
{\em Type of relation\/} & {\em Total\/} & {\em \# of cued relations\/}  \\ \hline \hline
Core:Contributor & 406 & 181\\ 
Joints & 64 & 19 \\
Clusters & 310 & 276 \\ \hline
Total & 780 & 476 \\
\hline \hline
\end{tabular}
\protect\caption{Distributions of relations and cue occurrences}
\label{distribution}
\end{table*}

An important result pointed out by \cite{MoserMooreACL95} is that cue
placement depends on core position.  When the core is first and
a cue is associated with the relation, the cue {\em never} occurs with
the core. In contrast, when the core is second, if a cue occurs, it
can occur either on the core or on the contributor.

\section{Learning from the corpus}

\label{learning}

\subsection{The algorithm}

We chose the C4.5 learning algorithm \cite{quinlan93} because it is
well suited to a domain such as ours with discrete valued  attributes.
Moreover, C4.5 produces decision trees and rule sets, both often used
in text generation to implement mappings from function features to
forms.\footnote{We will discuss only decision trees here.}  Finally,
C4.5 is both readily available, and is a benchmark learning algorithm
that has been extensively used in NLP applications, e.g.
\cite{litman-jair96,mooney96,neg-inlg96}.

As our dataset is small, the results we report are based on {\em
cross-validation}, which \cite{weiss-kul91} recommends as the best
method to evaluate decision trees on datasets whose cardinality is in
the hundreds.  Data for learning should be divided into {\em training}
and {\em test} sets; however, for small datasets this has the
disadvantage that a sizable portion of the data is not available for
learning. Cross-validation obviates this problem by running the
algorithm N times (N=10 is a typical value): in each run,
$\frac{(N-1)}{N}$th of the data, randomly chosen, is used as the {\em
training} set, and the remaining $\frac{1}{N}$th used as the {\em
test} set. The error rate of a tree obtained by using the whole
dataset for training is then assumed to be the average error rate on
the {\em test} set over the N runs. Further, as C4.5 prunes the
initial tree it obtains to avoid overfitting, it computes both {\em
actual} and {\em estimated} error rates for the pruned tree; see
\cite[Ch.~4]{quinlan93} for details. Thus, below we will report the
average {\em estimated} error rate on the test set, as computed by
10-fold cross-validation experiments.

\subsection{The features}

Each data point in our dataset corresponds to a {\em core:contributor}
relation, and is characterized by the following features, summarized in
Table~\ref{features}. 

\begin{table*}
\centering
\small
\begin{tabular}{||l||l|l||}\hline \hline
{\em feature type} & {\em feature\/} & {\em description\/} \\ \hline \hline
{\em Segment structure} 
& Trib-pos &  relative position of contrib in segment $+$\\
&  & number of contribs before and after core \\
& Inten-structure & intentional structure of segment \\ 
&Infor-structure & informational structure of segment \\ \hline 
{\em Core:contributor\/}  
& Inten-rel & enable, convince, concede \\
{\em relation\/} & Info-rel & 4 classes of  about 30 distinct relations \\
& Syn-rel &  independent sentences / segments,\\
& &  coordinated clauses, subordinated clauses \\ 
& Adjacency& are core and contributor adjacent? \\ \hline
{\em Embedding} & Core-type & segment, minimal unit \\
& Trib-type & segment, minimal unit \\
& Above~/~Below & number of relations hierarchically  \\
& & above / below current relation \\
\hline \hline
\end{tabular}
\protect\caption{Features}
\label{features}
\end{table*}

\paragraph{Segment Structure.} Three features capture the global
structure of the segment in which the current {\em core:contributor}
relation appears.
\begin{itemize}
\item {\em (Con)Trib(utor)-pos(ition)\/} captures the
position of a particular contributor within the larger segment in which it
occurs, and  encodes the structure of the segment in terms of how many 
contributors precede and follow the core. For example, 
contributor \xaref{seg2}{-D}
in Figure~\ref{relexpl} is labeled as B1A3-2after, as it is the second
contributor following the core in a segment with 1 contributor before and 3
after the core.
\item {\em Inten(tional)-structure\/} indicates which
contributors in the segment bear the same intentional
relations to the core.
\item {\em Infor(mational)-structure\/}. Similar to intentional structure, 
but applied to informational relations.
\end{itemize}
\paragraph{Core:contributor relation.} These features more
specifically characterize the current {\em core:contributor}
relation.
\begin{itemize}
\item {\em Inten(tional)-rel(ation)\/}. One of {\em concede, convince,
enable}.
\item {\em Infor(mational)-rel(ation)\/}. About 30  informational
relations have been coded for. 
However, as preliminary experiments showed that using them
individually results in overfitting the data, we classify them 
according to the four classes proposed in \cite{MoserMooreRDATR}:
{\em causality, similarity, elaboration, temporal\/}. 
{\em Temporal} relations only appear in clusters, thus not 
in the data we discuss in this paper.
\item {\em Syn(tactic)-rel(ation)\/}. Captures whether the core and
contributor are independent units (segments or sentences); whether
they are coordinated clauses; or  which of the two is
subordinate to the other.
\item {\em Adjacency}. Whether core and contributor are adjacent in
linear order. 
\end{itemize}
\paragraph{Embedding.} These features capture segment embedding,  {\em Core-type} and {\em Trib-type}
qualitatively,  and  {\em Above/Below} quantitatively.
\begin{itemize}
\item {\em Core-type/(Con)Trib(utor)-type}. Whether the core/the contributor
is a segment, or a minimal
unit (further subdivided into {\em action}, {\em state}, {\em matrix}).
\item {\em Above/Below}  
encode the number of relations hierarchically above and below the
current relation.
\end{itemize}

\subsection{The experiments}

Initially, we performed learning on all  406 instances of {\em core:contributor}
relations. We quickly determined that this approach would not lead to useful
decision trees. First, the trees we obtained were extremely complex (at least
50 nodes). Second, some of the subtrees corresponded to clearly identifiable
subclasses of the data, such as relations with an implicit core, which
suggested that we should  apply learning to these independently identifiable
subclasses.
Thus, we subdivided the data into three subsets: 
\begin{itemize}
\item {\em Core1\/}:  {\em core:contributor} relations with the core in first position 
\item {\em Core2}:  {\em core:contributor} relations with the core in second position
\item {\em Impl(icit)-core}:  {\em core:contributor} relations with an implicit core
\end{itemize}
While this has the disadvantage
of smaller training sets, the trees we obtain are  more manageable and more meaningful.
Table~\ref{dataset-distr} summarizes the cardinality of these sets, and the
frequencies of cue occurrence.

\begin{table*}
\centering
\begin{tabular}{||l||c|cc||}\hline \hline
{\em Dataset\/} & {\em \# of relations\/} & \multicolumn{2}{c||}{{\em \# of cued relations\/}} \\
\hline \hline
Core1  & 127 & \multicolumn{2}{c||}{52}\\ 
\hline
Core2 & 155 & \multicolumn{2}{c||}{100} \\
& & (on Trib: 43) & (on Core: 57) \\ 
\hline
Impl-core & 124 & \multicolumn{2}{c||}{29} \\ 
\hline \hline 
Total & 406 & \multicolumn{2}{c||}{181} \\
\hline \hline
\end{tabular}
\protect\caption{Distributions of relations and cue occurrences}
\label{dataset-distr}
\end{table*}

We ran four sets of experiments. In three of them we predict cue
occurrence and in one cue placement.\footnote{All our experiments are
run with {\em grouping} turned on, so that C4.5 groups values together
rather than creating a branch per value. The latter choice always
results in trees overfitted to the data in our domain. Using
classes of informational relations, rather than individual
informational relations, constitutes a sort of a priori grouping.}

\subsubsection{Cue Occurrence} 

Table~\ref{summary} summarizes our main results concerning cue
occurrence, and includes the error rates associated with different
feature sets. 
We adopt Litman's approach  \shortcite{litman-jair96} to determine
whether two error rates $\cal E$$_1$ and $\cal E$$_2$ are
significantly different. We compute 95\% confidence intervals for the
two error rates  using a {\em t\/}-test. $\cal E$$_1$ is
significantly better than $\cal E$$_2$ if the upper bound of the 95\%
confidence interval for $\cal E$$_1$ is lower than the lower bound of
the 95\% confidence interval for $\cal E$$_2$.

\begin{table*}
\centering
\small
\begin{tabular}{||l|l|l|l||}\hline \hline
& {\em Core1\/} & {\em Core2} &  {\em Impl-core\/} \\
\hline \hline
& & & \\
Baseline & 41.1 & 35.4 & 23.5 \\ \hline
& & & \\
Best  features & $\emptyset$ & Info-rel: 33.4$\pm$0.94   &  $\emptyset$  \\ \hline
& & & \\
Best tree & 25.6$\pm$1.24  (15) & 27.4$\pm$1.28  (18)  & 22.1$\pm$0.57
(10) \\ 
& & & \\
&  0. {\bf Trib-pos} & 0. {\bf Trib-Pos} &   0. Core-type \\ 
& 1. Trib-type & 1. Inten-rel &  1. {\bf Infor-struct}  \\
& 2. Syn-rel & 2. Info-rel &  2. Inten-rel \\
& 3. Core-type  & 3. Above &  \\
& 4. Above  & 4. Core-type  &   \\
& 5. Inten-rel & 5. Below  &  \\
\hline \hline
\end{tabular}
\protect\caption{Summary of learning results}
\label{summary}
\end{table*}

\begin{figure*}[tb]
\begin{center}
\mbox{\psfig{figure={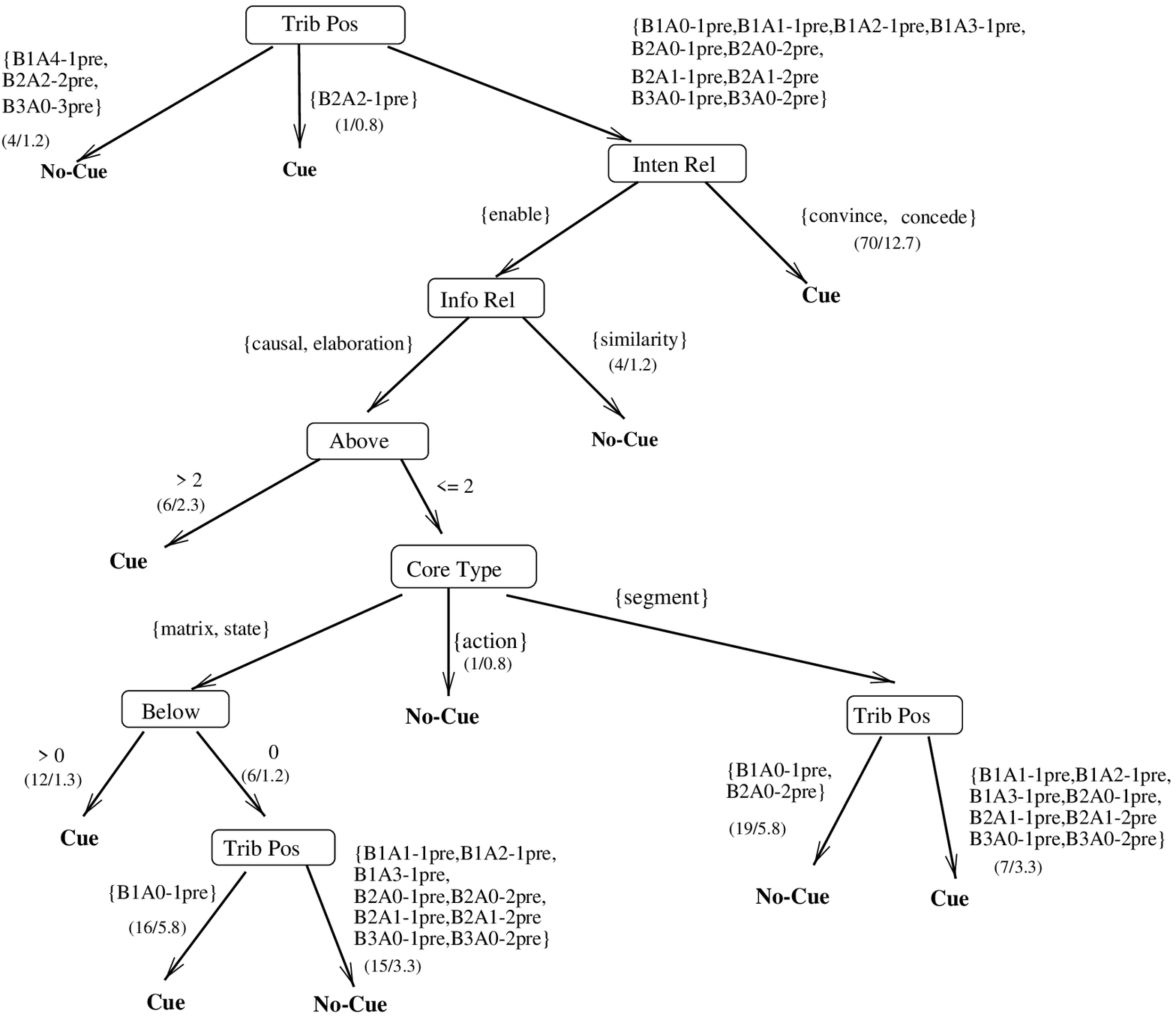},width=18cm,height=12cm,clip=-3cm}}
\end{center}
\caption{Decision tree for {\em Core2} --- occurrence}
\label{core2-tree}
\end{figure*}

\noindent
For each set of experiments, we report the following: 
\begin{enumerate}
\item  A baseline measure obtained
by choosing  
the majority class. E.g., for {\em Core1\/}  58.9\% of the relations are
not cued; thus, by deciding to   never include a cue,
one would be wrong 41.1\% of the times.
\item The  best individual features whose predictive power is better than 
the baseline: as Table~\ref{summary}  makes apparent, individual
features  do not have much predictive power. For neither
{\em Core1} nor {\em Impl-core} does any individual feature perform
better than the baseline, and for {\em Core2} only one feature is
sufficiently predictive. 
\item (One of) the best induced tree(s).  For each tree, we list the number of
nodes, and up to six of the  features that appear highest in the tree, with
their levels of embedding.\footnote{The trees that C4.5 generates are
right-branching, so this description is fairly adequate.} 
Figure~\ref{core2-tree} shows the tree for {\em Core2} (space
constraints prevent us from including figures for each
tree). 
In the figure, the  numbers in parentheses indicate  the number of cases
correctly covered by the leaf, and the number of expected errors at
that leaf.
\end{enumerate}

Learning turns out to be most useful for {\em Core1}, where the error
reduction (as percentage) from baseline to the upper bound of the best
result is 32\%; error reduction is 19\% for {\em Core2} and only 3\% for
{\em Impl-core}.

The best tree was obtained partly by informed choice, partly by trial
and error.
Automatically trying out all the
$2^{11}\:=\:2048$ subsets of features would be possible, but it would
require manual examination of about 2,000 sets of results, a daunting
task.  Thus, for each dataset we considered only the following subsets
of features.
\begin{enumerate}
\item All features. This always
results in C4.5 selecting a few features (from 3 to 7) for the final
tree.
\item   Subsets built out of  the 2 to 4 attributes appearing highest in the tree obtained by running
C4.5 on all features.
\item  In Table~\ref{features}, three features --- {\em Trib-pos, Inten-struct, 
Infor-struct} --- concern segment structure, eight do not. We
constructed three subsets by always including the eight features that
do not concern segment structure, and adding one of those that does.
The trees obtained by including {\em Trib-pos, Inten-struct,
Infor-struct} at the same time are in general more complex, and not
significantly better than other trees obtained by including only one
of these three features. We attribute this to the fact that these
features encode partly overlapping information.
\end{enumerate}

Finally, the best tree was obtained as follows.  We build the set of
trees that are statistically equivalent to the tree with the
best error rate (i.e., with the lowest error rate upper bound). Among
these trees, we choose the one that we deem the most perspicuous in
terms of features and of complexity. Namely, we pick the
simplest tree with {\em Trib-Pos} as the root if one exists, otherwise
the simplest tree.  Trees that have {\em Trib-Pos} as the root are the
most useful for text generation, because, given a complex segment,
{\em Trib-Pos} is the only attribute that unambiguously identifies
a specific contributor.  

Our results make apparent that the structure of segments plays a
fundamental role in determining cue occurrence.  One of the three
features concerning segment structure ({\em Trib-Pos, Inten-Structure,
Infor-Structure}) appears as the root or just below the root in all
trees in Table~\ref{summary}; more importantly, this same
configuration occurs in all  trees equivalent to the best tree
(even if the specific feature encoding segment structure may change). 
The level of embedding in a segment, as encoded by {\em Core-type},
{\em Trib-type}, {\em Above} and {\em Below} also figures
prominently. 

{\em Inten-rel} appears in all trees, confirming the
intuition that the speaker's purpose affects cue occurrence.  More
specifically, in Figure~\ref{core2-tree}, {\em Inten-rel}
distinguishes two different speaker purposes, {\em convince\/} and
{\em enable}. The same split occurs  in some of the best trees
induced on {\em Core1}, with the same outcome: i.e., {\em convince\/}
directly correlates with the occurrence of a cue, whereas for {\em
enable} other
features must be taken into account.\footnote{We can't draw any conclusions
concerning {\em concede}, as there are only 24 occurrences of {\em concede}
out of  406 {\em core:contributor} relations.}
Informational relations do not appear as often as intentional
relations; their discriminatory power seems more relevant for
clusters.  Preliminary experiments show that cue occurrence in
clusters depends only on informational and syntactic relations.
Finally, {\em Adjacency} does not seem to play any substantial role.

\subsubsection{Cue Placement}  

While cue occurrence and placement are  interrelated problems, we
performed learning on them separately. First, the issue of placement arises
only in the case of {\em Core2}; for {\em Core1}, cues {\em only}  occur
on the contributor.  Second, we attempted experiments on {\em
Core2} that discriminated between occurrence and placement at the same time, and
the derived trees were complex and not perspicuous. Thus, we ran an experiment on the
100 cued relations from {\em Core2} to investigate which factors
affect placing the  cue on the contributor in first position or on the core in
second; see Table~\ref{placement}.
\begin{table}
\centering
\small
\begin{tabular}{||l|l||}\hline \hline
Baseline & 43\% \\ \hline
Best  features &   Syn-rel: 24.1$\pm$0.69\\
& Trib-pos: 40$\pm$0.88  \\ \hline
Best tree &  20.6$\pm$0.97 (5)\\ 
&  0. Syn-rel \\ 
&  1.  Trib-pos\\
\hline \hline
\end{tabular}
\protect\caption{Cue placement on {\em Core2}}
\label{placement}
\end{table}

\begin{figure*}[tb]
\begin{center}
\mbox{\psfig{figure={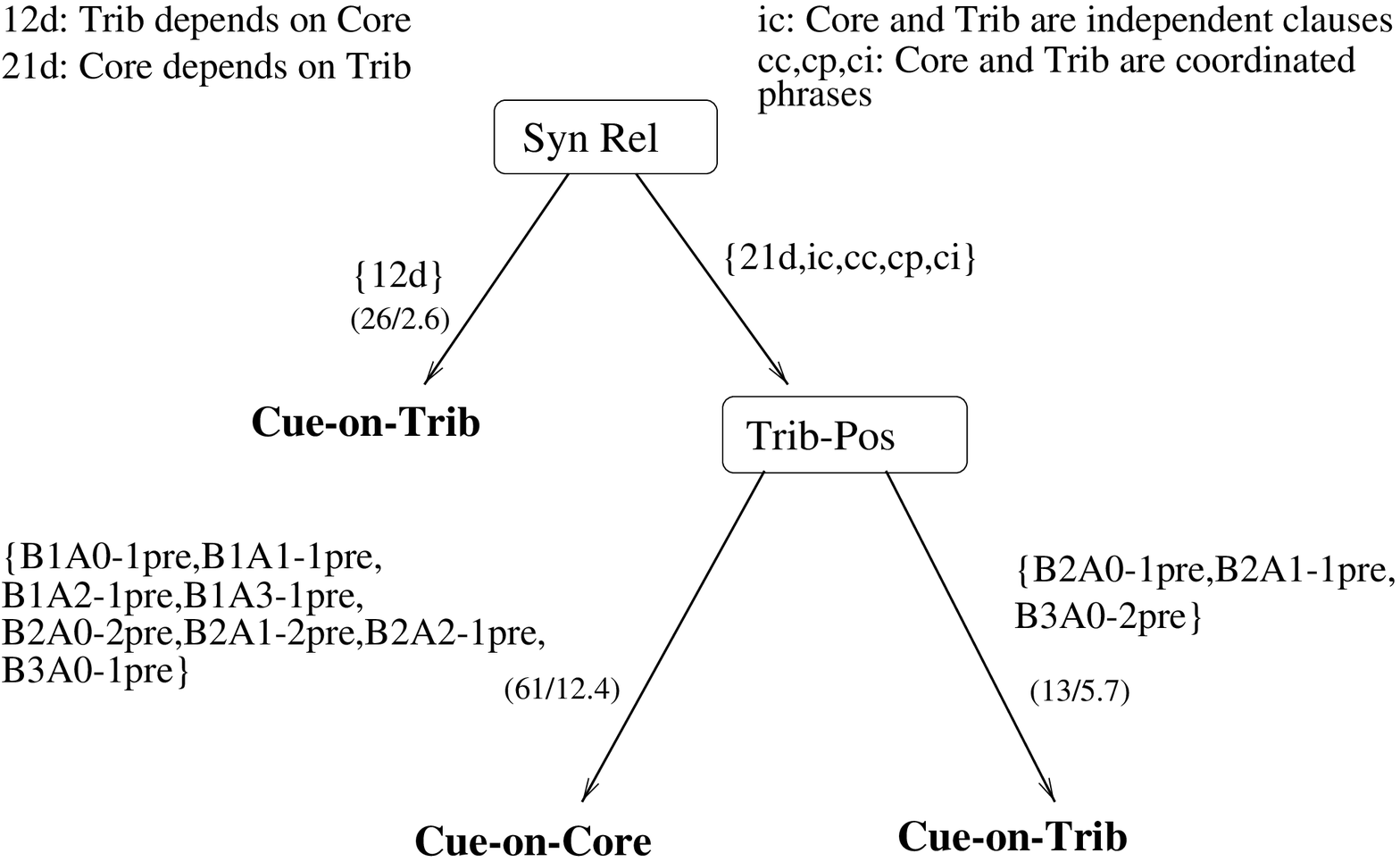},width=12cm,height=7cm,clip=-3cm}}
\end{center}
\caption{Decision tree for {\em Core2} --- placement}
\label{core2-plac-tree}
\end{figure*}

We ran the same trials discussed above on this dataset. In this case,
the best tree --- see Figure~\ref{core2-plac-tree} --- results from
combining the two best individual features, and reduces the error rate
by 50\%.  The most discriminant feature turns out to be the syntactic
relation between the contributor and the core. However, segment
structure still plays an important role, via {\em Trib-pos}.

While the importance of {\em Syn-rel\/} for placement seems clear, its
role concerning occurrence requires further exploration. It is
interesting to note that the tree induced on {\em Core1} --- the only
case in which {\em Syn-rel} is relevant for occurrence --- includes
the same distinction as in Figure~\ref{core2-plac-tree}: namely, if
the contributor depends on the core, the contributor must be marked,
otherwise other features have to be taken into account.
Scott and de Souza \shortcite{ScottEWNLG90} point out that 
``there is a strong correlation between the syntactic specification
of a complex sentence and its perceived rhetorical structure.''
It seems  that certain syntactic structures  function as a
cue.

\section{Discussion and Conclusions}

\label{conclude}

We have presented the results of machine learning experiments
concerning cue occurrence and placement.  As \cite{litman-jair96}
observes, this sort of empirical work supports the utility of machine
learning techniques applied to coded corpora.  As our study shows, 
individual features have no predictive power for cue
occurrence.
Moreover, it is hard to see how the best combination of individual features
could be found by manual inspection.

Our results also provide guidance for those building text
generation systems.  This study clearly indicates that segment
structure, most notably the ordering of core and contributor,
is crucial for determining cue occurrence.  Recall that 
it was only by considering {\em Core1} and {\em Core2} relations in 
distinct datasets that we were able to obtain perspicuous 
decision trees  that significantly reduce the error rate.  

This indicates that the representations produced by discourse planners
should distinguish those elements that constitute the core of each
discourse segment, in addition to representing the hierarchical
structure of segments.  Note that the notion of core is related to the
notions of {\em nucleus\/} in RST, {\em intended effect} in
\cite{dpocl}, and of {\em point} of a move in \cite{ElhadadConn}, and
that text generators representing these notions exist.

Moreover, in order to use the decision trees derived here,
decisions about whether or not to make the core explicit
and how to order the core and contributor(s) must be 
made before deciding cue occurrence, e.g., by exploiting 
other factors such as {\em focus} \cite{McKeownBook} and a discourse history.

Once decisions about {\em core:contributor} ordering and cue occurrence
have been made, a generator must still determine where to place cues
and select appropriate lexical items.  A major focus of our future
research is to explore the relationship between the selection and
placement decisions. Elsewhere, we have found that particular lexical
items tend to have a preferred location, defined in terms of
functional (i.e., core or contributor) and linear (i.e., first or
second relatum) criteria \cite{MoserMooreLanguage}.  Thus, if a
generator uses decision trees such as the one shown in
Figure~\ref{core2-plac-tree} to determine where a cue should be
placed, it can then select an appropriate cue from those that can mark
the given intentional / informational relations, and are usually
placed in that functional-linear location.  To evaluate this strategy,
we must do further work to understand whether there are important
distinctions among cues (e.g., {\em so, because}) apart from their
different preferred locations.  The work of Elhadad
\shortcite{ElhadadConn} and Knott \shortcite{KnottThesis} will help in
answering this question.

Future work comprises further probing into machine learning techniques,
in particular investigating whether  other learning algorithms
are more appropriate for our problem \cite{mooney96}, especially 
algorithms that  take into account some a priori knowledge
about features and their dependencies.

\section*{Acknowledgements}

This research is supported by the Office of Naval Research, Cognitive
and Neural Sciences Division (Grants N00014-91-J-1694 and
N00014-93-I-0812).  Thanks to Megan Moser for her prior work on this
project and for comments on this paper; to Erin Glendening and Liina
Pylkk\"{a}nen for their coding efforts; to Haiqin Wang for running many
experiments; to Giuseppe Carenini and Steffi Br\"{u}ninghaus for
discussions about machine learning.


\vspace*{-4ex}
\begin{thebibliography}{}

\bibitem[\protect\citename{Cohen}1984]{CohenColing84}
Cohen, Robin.
\newblock 1984.
\newblock A computational theory of the function of clue words in argument
  understanding.
\newblock In {\em Proceedings of COLING84}, pages 251--258, Stanford,
CA.

\bibitem[\protect\citename{Elhadad and McKeown}1990]{ElhadadConn}
Elhadad, Michael and Kathleen McKeown.
\newblock 1990.
\newblock Generating connectives.
\newblock In {\em Proceedings of COLING90}, pages 97--101, Helsinki,
Finland.

\bibitem[\protect\citename{Goldman}1988]{GoldmanTR}
Goldman, Susan~R.
\newblock 1988.
\newblock The role of sequence markers in reading and recall: Comparison of
  native and nonnative english speakers.
\newblock Technical report, University of California, Santa Barbara.

\bibitem[\protect\citename{Grosz and Sidner}1986]{GroszSidnerCL}
Grosz, Barbara~J. and Candace~L. Sidner.
\newblock 1986.
\newblock Attention, intention, and the structure of discourse.
\newblock {\em Computational Linguistics}, 12(3):175--204.

\bibitem[\protect\citename{Hobbs}1985]{Hobbs85}
Hobbs, Jerry~R.
\newblock 1985.
\newblock On the coherence and structure of discourse.
\newblock Technical Report CSLI-85-37, Center for the Study of Language and
  Information,  Stanford  University. 

\bibitem[\protect\citename{Knott}1996]{KnottThesis}
Knott, Alistair.
\newblock 1996.
\newblock {\em A Data-Driven methodology for motivating a set of coherence
  relations}.
\newblock {Ph.D.} thesis, University of Edinburgh.

\bibitem[\protect\citename{Litman}1996]{litman-jair96}
Litman, Diane~J.
\newblock 1996.
\newblock Cue phrase classification using machine learning.
\newblock {\em Journal of Artificial Intelligence Research}, 5:53--94.

\bibitem[\protect\citename{Litman and Allen}1987]{LitmanCogSci}
Litman, Diane~J. and James~F. Allen.
\newblock 1987.
\newblock A plan recognition model for subdialogues in conversations.
\newblock {\em Cognitive Science}, 11:163--200.

\bibitem[\protect\citename{Mann and Thompson}1986]{MannThompsonDP}
Mann, William~C. and Sandra~A. Thompson.
\newblock 1986.
\newblock Relational propositions in discourse.
\newblock {\em Discourse Processes}, 9:57--90.

\bibitem[\protect\citename{Mann and Thompson}1988]{MannRSTTEXT}
Mann, William~C. and Sandra~A. Thompson.
\newblock 1988.
\newblock {R}hetorical {S}tructure {T}heory: Towards a functional theory of
  text organization.
\newblock {\em TEXT}, 8(3):243--281.

\bibitem[\protect\citename{McKeown}1985]{McKeownBook}
McKeown, Kathleen~R.
\newblock 1985.
\newblock {\em Text Generation: Using Discourse Strategies and Focus
  Constraints to Generate Natural Language Text}.
\newblock Cambridge University Press, Cambridge, England.

\bibitem[\protect\citename{McKeown and Elhadad}1991]{McKeownElhadadNLGW88Book}
McKeown, Kathleen~R. and Michael Elhadad.
\newblock 1991.
\newblock A contrastive evaluation of functional unification grammar for
  surface language generation: A case study in the choice of connectives.
\newblock In C.~L. Paris, W.~R. Swartout, and W.~C. Mann,
  eds., {\em Natural Language Generation in Artificial Intelligence and
  Computational Linguistics}. Kluwer Academic Publishers, Boston, pages
  351--396.

\bibitem[\protect\citename{Millis, Graesser, and Haberlandt}1993]{Millisetal}
Millis, Keith, Arthur Graesser, and Karl Haberlandt.
\newblock 1993.
\newblock The impact of connectives on the memory for expository text.
\newblock {\em Applied Cognitive Psychology}, 7:317--339.

\bibitem[\protect\citename{Mooney}1996]{mooney96}
Mooney, Raymond~J.
\newblock 1996.
\newblock Comparative experiments on disambiguating word senses: An
  illustration of the role of bias in machine learning.
\newblock In {\em Conference on Empirical Methods in Natural Language
  Processing}.

\bibitem[\protect\citename{Moser and Moore}1995]{MoserMooreACL95}
Moser, Megan and Johanna~D. Moore.
\newblock 1995.
\newblock Investigating cue selection and placement in tutorial discourse.
\newblock In {\em Proceedings of ACL95}, pages 130--135, Boston, MA.



\bibitem[\protect\citename{Moser and Moore}1997]{MoserMooreLanguage}
Moser, Megan and Johanna~D. Moore.
\newblock 1997.
\newblock A corpus analysis of discourse cues and relational discourse
  structure.
\newblock {\em Submitted for publication}.

\bibitem[\protect\citename{Moser, Moore, and Glendening}1996]{MoserMooreRDATR}
Moser, Megan, Johanna~D. Moore, and Erin Glendening.
\newblock 1996.
\newblock {Instructions for Coding Explanations: {Identifying} Segments,
  Relations and Minimal Units}.
\newblock Technical Report 96-17, University of Pittsburgh, Department of
  Computer Science.


\bibitem[\protect\citename{Quinlan}1993]{quinlan93}
Quinlan, J.~Ross.
\newblock 1993.
\newblock {\em C4.5: {P}rograms for {M}achine {L}earning}.
\newblock Morgan Kaufmann.

\bibitem[\protect\citename{Reichman-Adar}1984]{ReichmanAIJ}
Reichman-Adar, Rachel.
\newblock 1984.
\newblock Extended person-machine interface.
\newblock {\em Artificial Intelligence}, 22(2):157--218.

\bibitem[\protect\citename{R\"osner and Stede}1992]{RosnerNLGW6}
R\"osner, Dietmar and Manfred Stede.
\newblock 1992.
\newblock Customizing {RST} for the automatic production of technical manuals.
\newblock In R.~Dale, E.~Hovy, D.~R\"osner, and O.~Stock, eds., {\em
  6th International Workshop on Natural Language
  Generation},   Springer-Verlag, Berlin, pages 199--215.

\bibitem[\protect\citename{Schiffrin}1987]{SchiffrinBook}
Schiffrin, Deborah.
\newblock 1987.
\newblock {\em Discourse Markers}.
\newblock Cambridge University Press, New York.

\bibitem[\protect\citename{Scott and {de Souza}}1990]{ScottEWNLG90}
Scott, Donia and Clarisse~Sieckenius {de Souza}.
\newblock 1990.
\newblock Getting the message across in {RST}-based text generation.
\newblock In R.~Dale, C.~Mellish, and M.~Zock, eds., {\em Current Research
  in Natural Language Generation}. Academic Press, New York, pages 47--73.

\bibitem[\protect\citename{Siegel and McKeown}1994]{siegel-aaai94}
Siegel, Eric~V. and Kathleen~R. McKeown.
\newblock 1994.
\newblock Emergent linguistic rules from inducing decision trees:
  Disambiguating discourse clue words.
\newblock In {\em  Proceedings of AAAI94}, pages 820--826.



\bibitem[\protect\citename{{Vander Linden} and D{i Eugenio}}1996]{neg-inlg96}
{Vander Linden}, Keith and Barbara D{i Eugenio}.
\newblock 1996.
\newblock Learning micro-planning rules for preventative expressions.
\newblock In {\em  8th International Workshop on Natural Language
  Generation}, Sussex, UK.

\bibitem[\protect\citename{{Vander Linden} and Martin}1995]{VanderLindenCL}
{Vander Linden}, Keith and James~H. Martin.
\newblock 1995.
\newblock Expressing rhetorical relations in instructional text: A case study
  of the purpose relation.
\newblock {\em Computational Linguistics}, 21(1):29--58.

\bibitem[\protect\citename{Weiss and Kulikowski}1991]{weiss-kul91}
Weiss, Sholom~M. and Casimir Kulikowski.
\newblock 1991.
\newblock {\em Computer Systems that learn: classification and prediction
  methods from statistics, neural nets, machine learning, and expert systems}.
\newblock Morgan Kaufmann.

\bibitem[\protect\citename{Young and Moore}1994]{dpocl}
Young, R.~Michael and Johanna~D. Moore.
\newblock 1994.
\newblock D{POCL}: {A} {P}rincipled {A}pproach to {D}iscourse {P}lanning.
\newblock In {\em 7th International Workshop on Natural Language
  Generation},  Kennebunkport, Maine.

\end{thebibliography}
\end{document}